\documentclass{article}

\usepackage[preprint]{neurips_2025}

\usepackage[utf8]{inputenc} 
\usepackage[T1]{fontenc}    
\usepackage{hyperref}       
\usepackage{url}            
\usepackage{booktabs}       
\usepackage{amsfonts}       
\usepackage{nicefrac}       
\usepackage{microtype}      
\usepackage[dvipsnames]{xcolor}
\usepackage{graphicx}
\usepackage{subfigure}
\usepackage{amsthm}
\usepackage{amssymb}
\usepackage{enumitem}
\usepackage{wrapfig}
\usepackage{dsfont}
\usepackage[ruled, noend]{algorithm2e}
\usepackage{mathrsfs}
\usepackage{multicol}
\usepackage{threeparttable}
\usepackage{multirow}
\usepackage{minitoc}
\usepackage{comment}
\usepackage{mathtools}

\usepackage{scalerel,stackengine}
\stackMath
\newcommand\reallywidehat[1]{%
\savestack{\tmpbox}{\stretchto{%
  \scaleto{%
    \scalerel*[\widthof{\ensuremath{#1}}]{\kern-.6pt\bigwedge\kern-.6pt}%
    {\rule[-\textheight/2]{1ex}{\textheight}}
  }{\textheight}%
}{0.5ex}}%
\stackon[1pt]{#1}{\tmpbox}%
}
\newcommand\reallywidecheck[1]{%
\savestack{\tmpbox}{\stretchto{%
  \scaleto{
    \scalerel*[\widthof{\ensuremath{#1}}]{\kern-.6pt\bigwedge\kern-.6pt}%
    {\rule[-\textheight/2]{1ex}{\textheight}}
  }{\textheight}%
}{0.5ex}}%
\stackon[1pt]{#1}{\scalebox{-1}{\tmpbox}}%
}

\theoremstyle{remark}

\newcommand{\qa}[1]{{\textcolor[HTML]{6491EA}{\textbf{#1}}}}
\newcommand{\mygreen}[1]{{\textcolor[HTML]{41BB94}{\textbf{#1}}}}
\newcommand{\myblue}[1]{{\textcolor[HTML]{5393D6}{\textbf{#1}}}}

\title{RealDrive: Retrieval-Augmented Driving \\ with Diffusion Models}

\author{%
  Wenhao Ding$^1$\ \ Sushant Veer$^1$\ \ Yuxiao Chen$^1$\ \ Yulong Cao$^1$\\ \textbf{Chaowei Xiao}$^{1,2}$\ \ \textbf{Marco Pavone}$^{1,3}$ \\
  $^1$NVIDIA Research \ \ \ $^2$University of Wisconsin -- Madison\ \ \ $^3$Stanford University \\
  \texttt{wenhaod@nvidia.com} \\
}

\begin{document}
\maketitle

\begin{abstract}
Learning-based planners generate natural human-like driving behaviors by learning to reason about nuanced interactions from data, overcoming the rigid behaviors that arise from rule-based planners. 
Nonetheless, data-driven approaches often struggle with rare, safety-critical scenarios and offer limited controllability over the generated trajectories. To address these challenges, we propose \textit{RealDrive}, a Retrieval-Augmented Generation (RAG) framework that initializes a diffusion-based planning policy by retrieving the most relevant expert demonstrations from the training dataset. By interpolating between current observations and retrieved examples through a denoising process, our approach enables fine-grained control and safe behavior across diverse scenarios, leveraging the strong prior provided by the retrieved scenario. Another key insight we produce is that a task-relevant retrieval model trained with planning-based objectives results in superior planning performance in our framework compared to a task-agnostic retriever.
Experimental results demonstrate improved generalization to long-tail events and enhanced trajectory diversity compared to standard learning-based planners -- we observe a 40$\%$ reduction in collision rate on the Waymo Open Motion dataset with RAG.
\end{abstract}

\section{Introduction}
\label{sec:intro}

\begin{wrapfigure}{R}{0.4\linewidth}
    \vspace{-4mm}
    \centering
    \includegraphics[width=\linewidth]{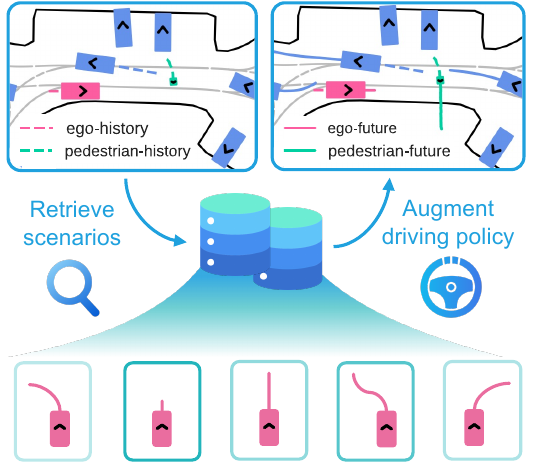}
    \vspace{-5mm}
    \caption{The ego vehicle (pink) first retrieves similar scenarios in the database, then uses the retrieved behaviors to augment the driving policy.}
    \label{fig:teaser}
    \vspace{-4mm}
\end{wrapfigure}

Recent advancements in autonomous driving have increasingly shifted motion planners towards learning-based~\cite{cheng2024pluto, zheng2025diffusion} and end-to-end~\cite{chitta2022transfuser, hu2023planning} driving policies that predict planning waypoints directly from vectorized or visual inputs. This trend has largely been fueled by the ease of developing nuanced driving behaviors using data, replacing the need for painstaking manual parameter tuning driven by engineering intuition in rule-based planners and the removal of information bottlenecks that arise from hand-designed module interfaces. Despite their promise, learning-based planners suffer from two key limitations: (i) they struggle with generalizing to under-represented (long-tail) scenarios in the training dataset~\cite{ding2023survey}, which in turn affects the overall safety and performance of the autonomous vehicle; (ii) learning-based planners offer limited controllability of the planned trajectories which is usually restricted to route information~\cite{zheng2025diffusion} (e.g., lane centerlines, goal positions) or coarse high-level commands~\cite{shao2024lmdrive, tian2024drivevlm} (e.g., go straight slowly, turn left).
This work aims to mitigate these issues by incorporating knowledge through a diffusion-based Retrieval-Augmented Generation (RAG) framework~\cite{gao2023retrieval}, which is illustrated in Figure~\ref{fig:teaser}.

Intuitively, RAG addresses under-represented scenarios by providing the policy with expert demonstrations from similar situations in the retrieval database. This grounds the policy in real-world expert behavior and offers a strong prior for decision-making. Furthermore, this prior -- supplied in the form of retrievals -- can be leveraged to exert greater control over the planned trajectories by selecting the retrieved samples for the desired characteristics; for example, selecting samples from a defensive expert can yield more defensive driving behaviors. RAG also improves the diversity of generated trajectories, effectively improving coverage of the feasible trajectory space.

In autonomous driving, the integration of RAG has largely been explored with large language models (LLMs)~\cite{yuan2024rag, chang2025driving}. Although general-purpose language descriptions and video embeddings that encode scene-level information are effective tools for retrieving similar scenarios, they do so at the expense of overlooking critical task-specific information (e.g., focusing on visual similarity of scenes instead of planning-relevant aspects, such as agent behaviors).
Additionally, the mechanism by which LLMs incorporate retrieved examples is typically confined to textual reasoning, whereas trajectory planning requires alignment between current and retrieved scenarios in a shared representation space. 

To address these challenges, instead of using RAG in conjunction with an LLM, we develop a task-specific embedding model trained with trajectory planning objectives, enabling the extraction of driving-relevant features. Retrieved trajectories are used as initial conditions for a diffusion model~\cite{ho2020denoising},
which refines them into the final trajectory through a denoising process. Although other classes of planners can also be enhanced with retrieval, we choose diffusion-based models as they provide an avenue for persistent injection of the observations and actions from the retrieved scenario during the denoising steps. In particular, to adapt the retrieved trajectory to the current scene and context and generate a reliable output trajectory, we introduce a retrieval interpolation module (RIM). RIM interpolates between the retrieved and current observations and actions by a coefficient that is adapted throughout the diffusion steps according to a sigmoidal scheduler. During inference, multiple retrieved examples can be incorporated to enhance trajectory diversity and robustness.

In summary, this work makes the following key contributions:
\vspace{-2mm}
\begin{itemize}[leftmargin=0.2in]
    \item We introduce a task-specific embedding model optimized for retrieval in planning, capturing crucial information, such as multi-agent interactions and critical objects, for driving policies. 
    \vspace{-1mm}
    \item We propose a novel RAG framework integrated with a diffusion model to improve learning-based planning performance and safety.
    \vspace{-1mm}
    \item We demonstrate the benefits of our method on real-world driving datasets, analyze key factors that influence its effectiveness, and offer insights for future research. 
\end{itemize}

\vspace{-4mm}
\section{Related Works}
\vspace{-1mm}

\subsection{Learning-based Driving Planner}

Learning-based planning methods~\cite{tampuu2020survey} are attracting increasing attention due to their adaptability and broader coverage compared to traditional rule-based approaches~\cite{bouchard2022rule}. In the early stages of deep learning, initial research~\cite{bojarski2016end, bansal2018chauffeurnet} investigated the use of Convolutional Neural Networks (CNNs) and Recurrent Neural Networks (RNNs) for training end-to-end driving planners. More recently, transformer-based policies have gained prominence in this field, driven by their success in sequence modeling within natural language processing~\cite{vaswani2017attention}. Transfuser~\cite{chitta2022transfuser} integrates RGB image and LiDAR data to train an effective planning model in the Carla simulator~\cite{dosovitskiy2017carla}. UniAD~\cite{hu2023planning} introduces a modular transformer-based architecture using image-only inputs, achieving significant improvements across multiple tasks on the nuScenes dataset~\cite{caesar2020nuscenes}. VAD~\cite{jiang2023vad} and its successor VAD v2~\cite{chen2024vadv2} streamline the model design by decoupling the map and motion networks for parallel processing. PARA-Drive~\cite{weng2024drive} advances this further by fully parallelizing the entire architecture, enhancing BEV feature extraction, and setting a new benchmark in performance. Although transformer-based policies provide an avenue for injecting vision-language and other high-level features, their ability to control the generation of the low-level trajectories is limited.

Diffusion models -- originally invented for image generation~\cite{ho2020denoising} -- provide a way to exert greater control over the generated trajectories while still benefiting from transformer-based encoders. Their ability to capture multi-modality and to follow instructions~\cite{rombach2022high} have led to their rapid adoption for autonomous driving. For instance, MotionDiffusion~\cite{jiang2023motiondiffuser} and CTG~\cite{zhong2023guided} employ diffusion models to enhance the diversity of traffic scenario generation. DiffusionDrive~\cite{liao2024diffusiondrive} introduces the Truncated Diffusion Model, which integrates multi-modal anchors to learn a diverse action distribution. Diffusion-ES~\cite{yang2024diffusion} demonstrates the model's capacity for zero-shot instruction following in planning tasks. Likewise, Diffusion Planner~\cite{zheng2025diffusion} achieves safe and adaptable planning by jointly modeling prediction. Our interest in diffusion policies in this paper arises from their ability to inject prior knowledge in the denoising process, which also makes them suitable for RAG.

\vspace{-1mm}
\subsection{Retrieval Augmented Generation (RAG)}
\vspace{-1mm}

RAG is a widely adopted approach in LLMs~\cite{gao2023retrieval}, leveraging external knowledge sources to improve the quality of generated responses. Owing to its effectiveness in utilizing information from training data, RAG has been extended to various domains, including reinforcement learning~\cite{goyal2022retrieval}, molecular generation~\cite{lee2024molecule, wang2024retrieval}, and robotic manipulation~\cite{kuang2024ram, memmel2024strap, guo2025srsa}. In particular, tasks involving physical interaction with the environment require not only static information retrieval but also the integration of sequential behaviors~\cite{memmel2024strap} and affordance~\cite{kuang2024ram}.

In the context of autonomous driving, RAG has been explored extensively, particularly in models employing multi-modal LLMs. RAG-Driver~\cite{yuan2024rag} employs in-context learning to enhance interpretability by mapping video and control signal embeddings into a unified retrieval space. RAC3~\cite{wang2024rac3} addresses hallucinations and weak real-world grounding by retrieving corner cases. Driving-RAG~\cite{chang2025driving} focuses on improving the embedding, retrieval, and application processes for driving scenarios. Another approach~\cite{cai2024driving} retrieves context-specific traffic rules and guidelines to guide vehicle behavior. Beyond traditional text similarity methods, recent studies have introduced advanced retrieval metrics such as Optimal Transport. For example, RALAD~\cite{zuo2025ralad} reduces the domain gap between real and simulated data using Optimal Transport~\cite{santambrogio2015optimal}, while RealGen~\cite{ding2024realgen} applies the Wasserstein distance to train an embedding model for generating controllable scenarios. Unlike the methods mentioned above, in this paper, we will use RAG with diffusion for planning in autonomous driving.

\vspace{-1mm}
\subsection{Diffusion with Retrieval Augmented Generation}
\vspace{-1mm}

While RAG is commonly employed alongside MLLMs, a growing body of research explores its integration with diffusion models across various domains. Blattmann et al.~\cite{blattmann2022retrieval} enhance image generation by enabling the model to condition on similar examples retrieved from a database, thereby improving both quality and diversity. RAPID~\cite{jiang2025rapid} leverages RAG to improve generation performance while reducing memory requirements, inference costs, and preserving privacy. Tan et al.~\cite{tan2024ragdiffusion} increases the fidelity of clothing image generation by incorporating external knowledge to mitigate structural distortions and hallucinations. ReMoDiffuse~\cite{zhang2023remodiffuse} improves text-driven 3D human motion synthesis by incorporating semantically and kinematically relevant motion samples during the denoising process. Liu et al.~\cite{liu2024retrieval} introduces a reference-guided diffusion model that utilizes retrieved examples to enhance the accuracy of time series forecasting.

The most closely related studies to our work use RAG within policy models for real-world interaction tasks. 
R2-Diff~\cite{oba2023r2} employs SDEdit~\cite{meng2021sdedit} to introduce noise into retrieved motion sequences derived from encoded images, followed by a denoising process to generate trajectories for manipulation tasks. A key limitation of SDEdit lies in the challenge of appropriately selecting the noise magnitude. To address this, READ~\cite{oba2024read} performs direct diffusion between retrieved and target motions, enabling a more efficient diffusion pathway and avoiding the need for long forward-reverse processes. Similarly, RAGDP~\cite{odonchimedragdp} introduces two variants of Diffusion Policy~\cite{chi2023diffusion} that utilize retrieved expert demonstrations. However, neither READ~\cite{oba2024read} nor RAGDP~\cite{odonchimedragdp} incorporates the observations of retrieved samples during the generation, leading to a limitation where the models replicate expert behavior without accounting for the relationship between the retrieved and current observations; RealDrive addresses this limitation in the prior literature.

\begin{figure}[t]
    \centering
    \includegraphics[width=1.0\linewidth]{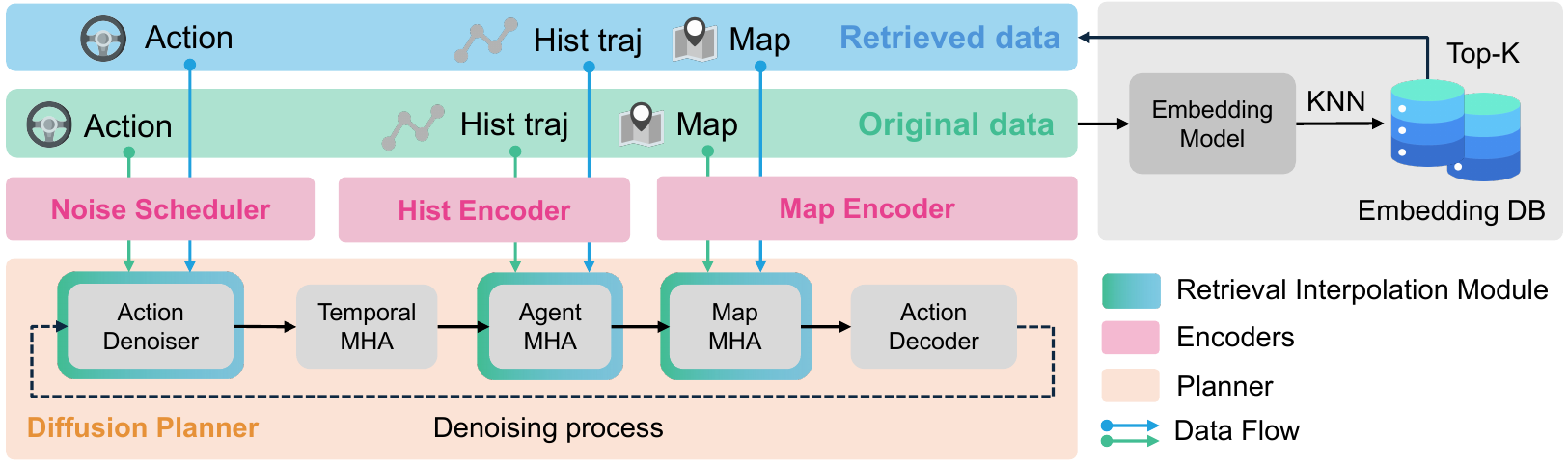}
    \vspace{-6mm}
    \caption{Training and inference pipeline of RealDrive. After sampling the \mygreen{original data}, the embedding model retrieves top-K similar scenarios from the database and uses them as the \myblue{retrieved data}. The historical trajectories and the map go through the encoders and are interpolated in the MHA. After going through the noise scheduler, the actions are interpolated in the action denoiser.}
    \label{fig:pipeline}
    \vspace{-5mm}
\end{figure}

\vspace{-1mm}
\section{Preliminary}
\vspace{-1mm}

In this section, we begin by formulating the planning problem addressed in this study, followed by an introduction to a basic diffusion-based planning model, which serves as the baseline.

\subsection{Problem Formulation}

A motion planning system utilizes the historical trajectories of surrounding agents, including the ego vehicle, in conjunction with the map context to generate multiple future trajectories for the ego vehicle.
The historical trajectory for an arbitrary timestep $t$ is denoted as $x^{[t-T_H:t]}_n$, where $T_H$ represents the number of past time steps and $n \in \{1,\cdots, N\}$ indexes the agents. The last point in history is used as the origin of the coordinate system. The predicted trajectory for the ego vehicle is represented as $x^{[t:t+T_F]}$, where $T_F$ is the prediction horizon. Each time step along the trajectory includes features such as position coordinates, heading, and velocity, which determine the state of an agent. 
The map context comprises any features that describe the map and its associated aspects, such as lane topology, wait lines, traffic lights, etc. However, for the experiments in this paper we will limit the map context to nearby lane centerlines, denoted by $m$ with dimensions $[S, P, 2]$, where $S$ is the number of lane segments, $P$ is the number of points per segment, and $2$ corresponds to the $x$ and $y$ coordinates. The observation at time $t$, denoted as $s_t:=(\{x^{[t-T_H:t]}_n\}_{n=1}^N,m)$, includes the historical trajectories up to time $t$ and the map context $m$.

To ensure the smoothness of the planning trajectory, we assume the agents to follow a kinematic bicycle model and use an inverse dynamics model to compute the acceleration and steering rate from the ground truth future trajectory and denote them as action $a_t$. During the training, the model outputs a sequence of actions $\{a_t\}$, which is then rolled out to trajectories using a bicycle model $\mathcal{F}$ (defined in the Appendix~\ref{app:dynamics}) to calculate the loss function.

\subsection{Diffusion Driving Planner}
\label{sec:diffusion_planner}

Diffusion models~\cite{ho2020denoising} represent a class of likelihood-based generative models that learn to reverse a predefined Markovian noise process, which incrementally perturbs data with Gaussian noise. In the forward process, an initial action sample $a_0$ is transformed over $H$ timesteps according to:
\begin{equation}
    q(a_h | a_{h-1}) := \mathcal{N}(a_h; \sqrt{1 - \beta_h} a_{h-1}, \beta_h \mathbf{I}),\ \ 
    q(a_h | a_0) := \mathcal{N}\left(a_h; \sqrt{\bar{\alpha}_h} a_0, (1- \bar{\alpha}_h)\mathbf{I}\right),
\label{equ:transition}
\end{equation}
where $\beta_h\in[0,1]$ forms a predefined noise schedule $\{\beta_h\}_{h=1}^{H}$, and $\bar{\alpha}_h = \prod_{i=1}^h(1-\beta_i)$. The reverse (generative) process is parameterized by a neural network $f_\theta$, which estimates the original data $\hat{a}_0 = f_\theta(a_h, h)$ from its corrupted counterpart. Training involves minimizing the expected mean squared error between the true and predicted clean trajectory from the forward dynamics model $\mathcal{F}$:
\begin{equation}
    \mathcal{L}_\theta := \underset{(h,\epsilon)\ \sim\ \mathcal{U}(\{1,\cdots,H\}) \times \mathcal{N}(0,\mathbf{I})}{\mathbb{E}} \|\mathcal{F}(a_0) - f_\theta\left[\sqrt{\bar{\alpha}_h} \mathcal{F}(a_0) + \sqrt{1- \bar{\alpha}_h}\epsilon, h\right] \|^2.
\label{equ:loss_function}
\end{equation}
During inference, the process begins with a sample $a_H\sim \mathcal{N}(0, \mathbf{I})$, and the denoising network $f_\theta$ is applied iteratively to obtain a sequence $\{a_{h-1}\}$ that converges to a high-quality data sample:
\begin{equation}
    p_{\theta}(a_{h-1}|a_{h}) := \mathcal{N}(a_{h-1}; \mu_h, \sigma_h^2 \mathbf{I}),
\label{equ:inference}
\end{equation}
where the mean $\mu_h$ and variance $\sigma_h^2$ are given by:
\begin{equation}
    \mu_h = \frac{\sqrt{\bar{\alpha}_{h-1}}\cdot \beta_h}{1-\bar{\alpha}_{h}} \cdot f_\theta(a_h, h) + \frac{\sqrt{1-\beta_h}\cdot (1-\bar{\alpha}_{h-1})}{1-\bar{\alpha}_{h}}\cdot a_h,\ \ \ \sigma_h^2 = \frac{1-\bar{\alpha}_{h-1}}{1-\bar{\alpha}_{h}}\cdot \beta_h.
\end{equation}

In this study, the diffusion model serves as the planning model, as illustrated in Figure~\ref{fig:pipeline}. We adopt the Diffusion Transformer architecture~\cite{peebles2023scalable} as the backbone, utilizing two transformer-based encoders to extract history and map features. For sampling, we employ the DDPM method~\cite{ho2020denoising} with a cosine noise schedule, following common practices in prior works~\cite{ajay2022conditional, chi2023diffusion}.

\vspace{-2mm}
\section{Retrieval-Augmented Driving Policy}
\vspace{-1mm}

The proposed RealDrive comprises two key components: an embedding model for scenario retrieval and an augmentation model that utilizes retrieved examples to guide planning. The following sections detail the design and functionality of both components.

\subsection{Embedding Model for Retrieval}

While prior work~\cite{ding2024realgen} has investigated embedding models to assess scenario similarity, it primarily focuses on the interaction between agents rather than decision-relevant information specific to the ego vehicle’s driving task. To address this limitation, we propose leveraging embeddings derived from a planning model to more effectively capture task-specific scenario similarities.

Specifically, we adopt the same architecture as the denoising model $f_\theta$ from Section~\ref{sec:diffusion_planner}, but modify it to serve as the embedding model $\mathcal{E}$. The input timestep $h$ is removed, and the original noisy data input is replaced with a learnable query embedding. The final transformer module produces an output of shape $[T_F, h_e]$, where $h_e$ denotes the embedding dimension. We compute the mean across all temporal dimensions, resulting in a fixed-size vector of shape $[h_e]$ used for similarity measurement. Scenario retrieval is performed using the Euclidean distance between embeddings to identify the top-K most similar cases. For efficient nearest neighbor search, we employ the FAISS library~\cite{johnson2019billion}, which supports multiple fast and approximate retrieval algorithms.

\newcommand\mycommfont[1]{\small\ttfamily\textcolor{RoyalBlue}{#1}}
\begin{figure}[t]
\centering
\begin{minipage}[t]{0.48\textwidth}
\begin{algorithm}[H]
\caption{RealDrive Training}
\label{algo:training}
\KwIn{Training dataset $\mathcal{D}$, Retrieval dataset $\mathcal{D}_r$, Embedding Model $\mathcal{E}$}
\KwOut{The denoise model $f_{\theta}$}
\For{$(s, a) \in \mathcal{D}$}{
    Retrieve $(s_r, a_r) \leftarrow \mathcal{E}(s, \mathcal{D}_r)$\;
    Sample $h\sim \mathcal{U}(\{1,\cdots,H\})$\;
    Get noisy action $\tilde{a}$ and $\tilde{a}_r$ using (\ref{equ:transition})\;
    $\hat{a} \leftarrow f_{\theta}(s, \tilde{a}, s_r, \tilde{a}_r, h)$ with (\ref{eq:obs_inter}) and (\ref{eq:action_inter})\;
    Roll out $\hat{x}^{t:t+T_F}$ from $\hat{a}$ with dynamics\;
    Update $\theta$ with $\mathcal{L}_\theta$ in (\ref{equ:loss_function})\;
}
\end{algorithm}
\end{minipage}
\hfill
\begin{minipage}[t]{0.5\textwidth}
\begin{algorithm}[H]
\caption{RealDrive Inference}
\label{algo:inference}
\KwIn{Retrieval dataset $\mathcal{D}_r$, Embedding Model $\mathcal{E}$, The denoise model $f_{\theta}$}
\KwOut{Generated action sequence $\hat{x}^{0:T_F}$}
Retrieve $(s_r, a_r) \leftarrow \mathcal{E}(s, \mathcal{D}_r)$\;
Initialize $\tilde{a} \sim \mathcal{N}(0, \mathbf{I})$\;
\For{$h=H$ \KwTo $1$}{
    Get noisy action $\tilde{a}_r$ using (\ref{equ:transition})\;
    $\hat{a} \leftarrow f_{\theta}(s, \tilde{a}, s_r, \tilde{a}_r, h)$ with (\ref{eq:obs_inter}) and (\ref{eq:action_inter})\;
    Update $\tilde{a}$ with $\hat{a}$ using (\ref{equ:inference})\;
}
Roll out $\hat{x}^{t:t+T_F}$ from $\hat{a}$ with dynamics\;
\end{algorithm}
\end{minipage}
\vspace{-4mm}
\end{figure}

\subsection{Augmentation via Retrieval Interpolation Module}

The central idea of RealDrive is to leverage retrieved scenarios as informative priors and progressively adapt their knowledge to the current driving context. Prior methods~\cite{oba2024read, odonchimedragdp} typically transfer only the action from retrieved examples, resulting in models that naively replicate retrieved behaviors without accounting for differences in contextual observations. This approach overlooks the relationship between retrieved and current scenarios, potentially impairing generalization.
To address this limitation, we introduce the Retrieval Interpolation Module (RIM), which integrates an implicit reasoning mechanism into the diffusion denoising process by jointly interpolating both observations and actions between the retrieved and current scenarios.

We define an interpolation coefficient $\lambda \in [0, 1]$ that governs the degree of transfer from the retrieved scenario. Since the rate of knowledge transfer should align with the denoising progression, we construct a flexible interpolation scheduler correlated with the denoising timestep. Specifically, we define $\lambda$ using a sigmoid-based heuristic:
\begin{equation}
    \lambda(\hat{h}) = S(\hat{h}; n, m) = \frac{\hat{h}^n}{\hat{h}^n + (1-\hat{h})^m},\ \ \ \hat{h} = \frac{h-1}{H-1}\in [0, 1],
\label{equ:sigmoid}
\end{equation}
where $n$ and $m$ are hyperparameters that control the curvature near $\hat{h}=0$ and $\hat{h}=1$, respectively.

For observation interpolation, we operate on the output embeddings of the multi-head attention (MHA) layers in the transformer blocks, following prior work demonstrating this strategy's controllability benefits~\cite{fan2024refdrop}. Given a learnable query $q$, a key embedding $k$, and its retrieved counterpart $k_r$, the interpolated output of the original MHA is replaced by the following:
\begin{equation}
    RIM(q, k, k_r) = (1 - \lambda) \cdot \text{MHA}(q, k, k) + \lambda \cdot \text{MHA}(q, k_r, k_r).
\label{eq:obs_inter}
\end{equation}
This latent-space interpolation supports flexible integration of diverse contextual features without explicit alignment of vehicle counts or lane structures across scenarios.

For action interpolation, we operate in the latent space after the denoising step, where both the current and retrieved noisy actions, denoted $\tilde{a}$ and $\tilde{a}_r$, are processed by a shared multi-layer perceptron (MLP). The interpolated result that will be used as the query embedding in MHA is computed as:
\begin{equation}
    q = RIM(\tilde{a}, \tilde{a}_r) = (1 - \lambda) \cdot \text{MLP}(\tilde{a}) + \lambda \cdot \text{MLP}(\tilde{a}_r),
\label{eq:action_inter}
\end{equation}
where $\tilde{a}$ and $\tilde{a}_r$ are sampled from the forward process $q(a_h|a_0)$ as described in Equation~(\ref{equ:transition}). Performing interpolation in the latent space ensures smooth integration without disrupting the underlying dynamics or spatial semantics.

\vspace{-1mm}
\section{Experiments}
\vspace{-1mm}

In this section, we first describe the experimental setup, followed by a presentation of results addressing the following research questions: \qa{Q1}: How effective is the retrieval model in identifying relevant scenarios? \qa{Q2}: How does RAG improve planning performance? \qa{Q3}: What factors influence the effectiveness of the RAG framework? \qa{Q4}: How does RAG impact inference efficiency?

\subsection{Experimental Setup}
\label{subsec:setting}

\textbf{Dataset.}
We conduct experiments on the nuScenes~\cite{caesar2020nuscenes} and Waymo Open Motion Dataset~\cite{ettinger2021large}. The nuScenes dataset includes 700 training scenes and 150 validation scenes, while Waymo comprises 487,004 training scenes and 44,097 validation scenes. Each scene is segmented into clips with a 2-second observation history and a 4-second prediction horizon at a frequency of 10 Hz. The retrieval database is constructed from the training data, with segments from the same scene excluded during retrieval to prevent information leakage.

\textbf{Baselines.}
To evaluate the impact of RAG, we compare it against a diffusion-based model that omits retrieval augmentation, denoted as w/o RAG. Additionally, we include two alternative inference settings: w/ RAG (Random), where randomly selected samples from the retrieval database are used for augmentation; and w/ RAG (Oracle), where segments augmented with ground-truth future trajectories are retrieved, providing an upper performance bound. To streamline the exposition, we assign the experimental setups a setting index as denoted in the second row of Tables~\ref{tab:nuscenes_results} and \ref{tab:waymo_results}.

\textbf{Metrics.}
We adopt open-loop planning metrics for quantitative evaluation. Specifically, minADE$_6$ and minFDE$_6$ represent the minimum average and final displacement errors, respectively, between the ground-truth trajectory and the $6$ predicted trajectories. The Time to Closest Encounter (TTCE)~\cite{eggert2014predictive} measures the time until the minimum distance between the ego vehicle and surrounding agents, irrespective of collision. The Collision Rate (CR) quantifies the average number of collisions per segment across the $6$ generated trajectories.

\begin{figure}[t]
    \centering
    \includegraphics[width=1.0\linewidth]{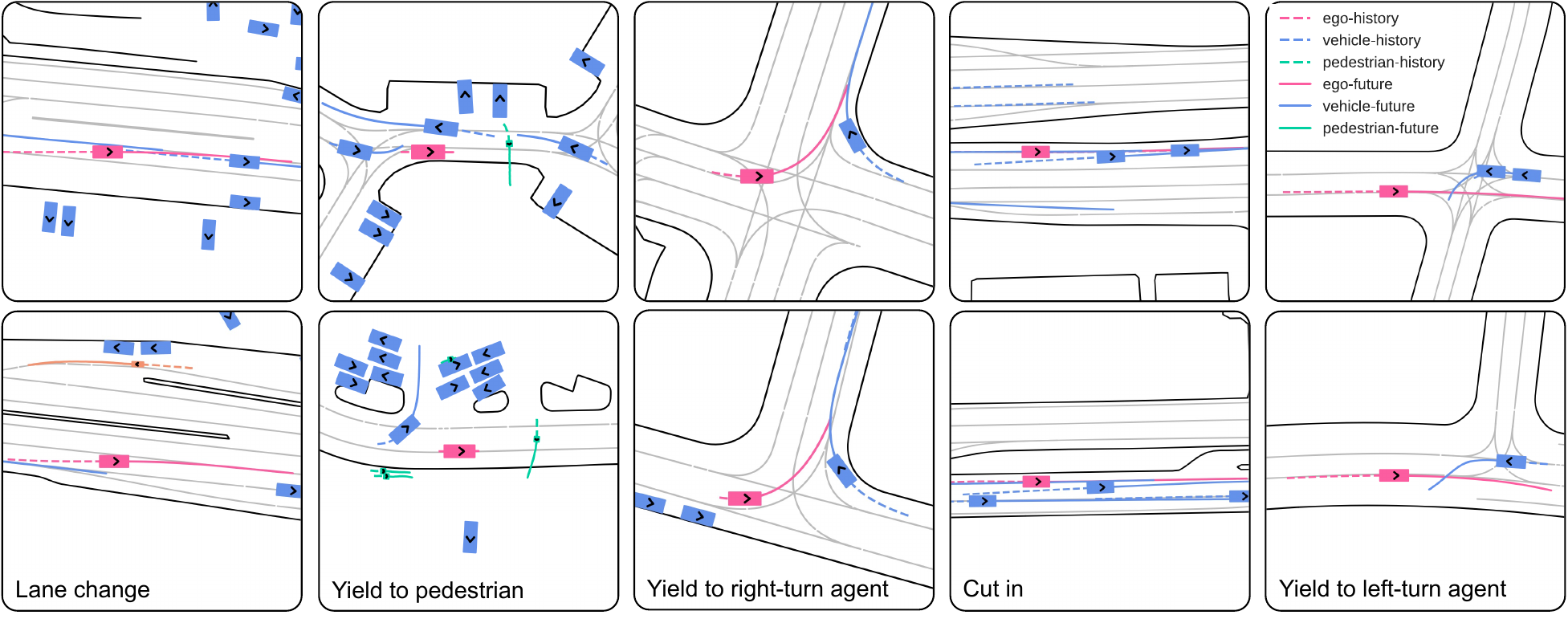}
    \vspace{-6mm}
    \caption{Examples of retrieved scenarios. The top row shows the query scenarios, and the bottom row shows the top-1 retrieved scenarios using our task-specific embedding model.}
    \label{fig:retrieval}
    \vspace{-5mm}
\end{figure}

\begin{table}[ht]
\caption{Open-loop planning evaluation results (mean$\pm$std) on nuScenes dataset}
\label{tab:nuscenes_results}
\vspace{-1mm}
\centering
\resizebox{1.0\linewidth}{!}{
\begin{threeparttable}
\begin{tabular}{l|c@{\hspace{5pt}}c|c@{\hspace{5pt}}c|c@{\hspace{5pt}}c|c@{\hspace{5pt}}c}
\toprule
Inference  & \multicolumn{2}{c|}{w/o RAG} & \multicolumn{2}{c|}{w/ RAG} & \multicolumn{2}{c|}{w/ RAG (Random)} & \multicolumn{2}{c}{w/ RAG (Oracle)} \\
Training   & w/o RAG & w/ RAG & w/o RAG & w/ RAG$^\dag$ & w/o RAG & w/ RAG & w/o RAG & w/ RAG \\
\midrule
Setting Index & 1 & 2 & 3 & 4 & 5 & 6 & 7 & 8 \\
\midrule
minADE$_6$ ($\downarrow$) 
& 0.527\scriptsize{$\pm .005$} & 0.542\scriptsize{$\pm .004$} & 0.570\scriptsize{$\pm .003$} & \textbf{0.497}\scriptsize{$\pm .002$} 
& 0.574\scriptsize{$\pm .006$} & 0.536\scriptsize{$\pm .004$} & 0.553\scriptsize{$\pm .005$} & \underline{0.487}\scriptsize{$\pm .003$} \\
minFDE$_6$ ($\downarrow$) 
& 1.583\scriptsize{$\pm .012$} & 1.582\scriptsize{$\pm .015$} & 1.733\scriptsize{$\pm .009$} & \textbf{1.476}\scriptsize{$\pm .009$} 
& 1.811\scriptsize{$\pm .010$} & 1.669\scriptsize{$\pm .008$} & 1.686\scriptsize{$\pm .013$} & \underline{1.460}\scriptsize{$\pm .011$} \\
minCR ($\downarrow$)      
& 0.012\scriptsize{$\pm .000$} & 0.019\scriptsize{$\pm .001$} & 0.014\scriptsize{$\pm .000$} & \underline{\textbf{0.011}}\scriptsize{$\pm .000$} 
& 0.012\scriptsize{$\pm .000$} & 0.019\scriptsize{$\pm .001$} & 0.013\scriptsize{$\pm .000$} & 0.012\scriptsize{$\pm .000$} \\
avgCR ($\downarrow$)      
& 0.031\scriptsize{$\pm .002$} & 0.045\scriptsize{$\pm .002$} & 0.026\scriptsize{$\pm .001$} & \underline{\textbf{0.020}}\scriptsize{$\pm .002$} 
& 0.049\scriptsize{$\pm .001$} & 0.064\scriptsize{$\pm .003$} & 0.027\scriptsize{$\pm .001$} & 0.031\scriptsize{$\pm .002$} \\
minTTCE ($\uparrow$)      
& 0.119\scriptsize{$\pm .003$} & 0.119\scriptsize{$\pm .002$} & 0.124\scriptsize{$\pm .003$} & \textbf{0.131}\scriptsize{$\pm .002$} 
& 0.114\scriptsize{$\pm .004$} & 0.109\scriptsize{$\pm .004$} & 0.130\scriptsize{$\pm .002$} & \underline{0.134}\scriptsize{$\pm .001$} \\
avgTTCE ($\uparrow$)      
& 0.155\scriptsize{$\pm .003$} & 0.156\scriptsize{$\pm .004$} & 0.158\scriptsize{$\pm .002$} & \textbf{0.161}\scriptsize{$\pm .004$} 
& 0.157\scriptsize{$\pm .004$} & 0.161\scriptsize{$\pm .004$} & 0.161\scriptsize{$\pm .001$} & \underline{0.165}\scriptsize{$\pm .002$}  \\
\bottomrule
\end{tabular}
\begin{tablenotes}
\footnotesize
\item[*]Underline means the best result. Bold font means the best result, excluding settings 7 and 8. Numbers are averaged over 5 runs.
\item[$\dag$]The setting of RealDrive.
\end{tablenotes}
\end{threeparttable}
}
\vspace{-2mm}
\end{table}

\begin{table}[ht]
\caption{Open-loop planning evaluation results (mean$\pm$std) on Waymo dataset}
\label{tab:waymo_results}
\vspace{-1mm}
\centering
\resizebox{1.0\linewidth}{!}{
\begin{threeparttable}
\begin{tabular}{l|c@{\hspace{5pt}}c|c@{\hspace{5pt}}c|c@{\hspace{5pt}}c|c@{\hspace{5pt}}c}
\toprule
Inference  & \multicolumn{2}{c|}{w/o RAG} & \multicolumn{2}{c|}{w/ RAG} & \multicolumn{2}{c|}{w/ RAG (Random)} & \multicolumn{2}{c}{w/ RAG (Oracle)} \\
Training   & w/o RAG & w/ RAG & w/o RAG & w/ RAG$^\dag$ & w/o RAG & w/ RAG & w/o RAG & w/ RAG \\
\midrule
Setting Index & 1 & 2 & 3 & 4 & 5 & 6 & 7 & 8 \\
\midrule
minADE$_6$ ($\downarrow$) 
& 0.164\scriptsize{$\pm .002$} & 0.163\scriptsize{$\pm .001$} & 0.168\scriptsize{$\pm .001$} & \textbf{0.152}\scriptsize{$\pm .001$} 
& 0.498\scriptsize{$\pm .004$} & 0.473\scriptsize{$\pm .003$} & 0.097\scriptsize{$\pm .000$} & \underline{0.069}\scriptsize{$\pm .000$} \\
minFDE$_6$ ($\downarrow$) 
& 0.597\scriptsize{$\pm .003$} & 0.593\scriptsize{$\pm .004$} & 0.626\scriptsize{$\pm .004$} & \textbf{0.544}\scriptsize{$\pm .005$} 
& 1.703\scriptsize{$\pm .012$} & 1.640\scriptsize{$\pm .015$} & 0.347\scriptsize{$\pm .003$} & \underline{0.230}\scriptsize{$\pm .002$} \\
minCR ($\downarrow$)      
& 0.0016\scriptsize{$\pm .00$} & 0.0014\scriptsize{$\pm .00$} & 0.0016\scriptsize{$\pm .00$} & \textbf{0.0012}\scriptsize{$\pm .00$} 
& 0.0016\scriptsize{$\pm .00$} & 0.0015\scriptsize{$\pm .00$} & 0.0014\scriptsize{$\pm .00$} & \underline{0.0011}\scriptsize{$\pm .00$} \\
avgCR ($\downarrow$)      
& 0.0043\scriptsize{$\pm .00$} & 0.0046\scriptsize{$\pm .00$} & 0.0032\scriptsize{$\pm .00$} & \textbf{0.0026}\scriptsize{$\pm .00$} 
& 0.0421\scriptsize{$\pm .00$} & 0.0401\scriptsize{$\pm .00$} & 0.0029\scriptsize{$\pm .00$} & \underline{0.0021}\scriptsize{$\pm .00$} \\
minTTCE ($\uparrow$)      
& 0.083\scriptsize{$\pm .001$} & 0.088\scriptsize{$\pm .000$} & 0.091\scriptsize{$\pm .001$} & \textbf{0.095}\scriptsize{$\pm .001$} & 0.065\scriptsize{$\pm .001$} & 0.062\scriptsize{$\pm .001$} & 0.095\scriptsize{$\pm .001$} & \underline{0.096}\scriptsize{$\pm .000$} \\
avgTTCE ($\uparrow$)      
& 0.107\scriptsize{$\pm .001$} & 0.108\scriptsize{$\pm .001$} & 0.108\scriptsize{$\pm .000$} & \textbf{0.112}\scriptsize{$\pm .001$} & 0.103\scriptsize{$\pm .002$} & 0.102\scriptsize{$\pm .002$} & 0.110\scriptsize{$\pm .001$} & \underline{0.114}\scriptsize{$\pm .001$} \\
\bottomrule
\end{tabular}
\begin{tablenotes}
\footnotesize
\item[*]Underline means the best result. Bold font means the best result, excluding settings 7 and 8. Numbers are averaged over 5 runs.
\item[$\dag$]The setting of RealDrive.
\end{tablenotes}
\end{threeparttable}
}
\vspace{-4mm}
\end{table}

\subsection{Effectiveness of the Retrieval Embedding Model (\qa{Q1})}

As the foundation of RealDrive, the quality of the retrieval embedding model plays a critical role in the overall system performance. To assess its effectiveness, we conduct a qualitative analysis by visualizing both the query scenario and its top-1 retrieved counterpart, as shown in Figure~\ref{fig:retrieval}. The retrieved examples consistently demonstrate strong semantic alignment with the query, capturing not only straightforward behaviors but also complex traffic dynamics such as single-agent maneuvers (e.g., cut-ins) and multi-agent interactions (e.g., yielding to a left-turning vehicle). These observations indicate that the embedding model successfully encodes task-relevant scenario features.

\subsection{Impact of RAG on Planning Performance (\qa{Q2})}

\begin{wrapfigure}{R}{0.4\linewidth}
    \vspace{-4.5mm}
    \centering
    \includegraphics[width=1.0\linewidth]{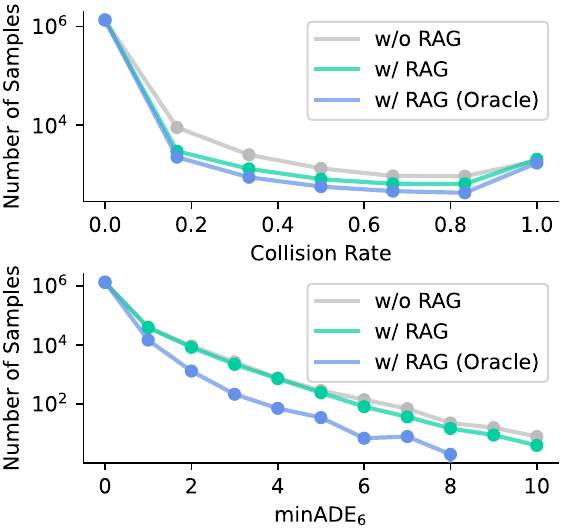}
    \vspace{-7mm}
    \caption{Using RAG reduces the number of samples with high collision rate and minade$_6$. Using RAG with oracle samples can further reduce the number.}
    \vspace{-4mm}
    \label{fig:dist}
\end{wrapfigure}

We evaluate the influence of retrieval augmentation on planning performance using the nuScenes and Waymo datasets, with results summarized in Table~\ref{tab:nuscenes_results} and Table~\ref{tab:waymo_results}, respectively. From these results, several consistent trends emerge:
\textbf{RAG Enhances Performance:} Comparing Setting 1 and Setting 4, we observe that retrieval augmentation leads to improvements in both trajectory accuracy and safety metrics.
\textbf{Importance of Joint Training:} Setting 3 performs worse than Setting 4, underscoring the necessity of training the model jointly with retrieval augmentation to effectively utilize the retrieved information.
\textbf{Non-Disruptive Training:} Setting 2 performs comparably to the baseline, indicating that incorporating RAG during training does not degrade performance in inference w/o RAG.
\textbf{Benefit of High-Quality Retrieval:} Setting 8, which uses oracle retrieval with ground-truth future information, achieves the best results across most metrics, demonstrating the upper bound potential of retrieval-based methods.
\textbf{Sensitivity to Retrieval Quality:} Settings 5 and 6 highlight that poor-quality retrievals can significantly degrade performance, emphasizing the importance of a well-trained embedding model.

\textbf{Performance in Long-Tail Scenarios.}
While RAG provides an overall performance gain, its effectiveness truly shines in handling under-represented scenarios in the training data. Figure~\ref{fig:dist} presents the distributions of minADE$_6$ and Collision Rate for the three inference settings described in Table~\ref{tab:waymo_results}. The results reveal that RAG primarily improves performance in the tail of the distribution, significantly reducing both collision frequency and trajectory error in rare, safety-critical cases. These findings support our initial hypothesis in Section~\ref{sec:intro}, which posits that RAG is particularly beneficial in handling under-represented scenarios in the training data.

\begin{figure}[t]
    \centering
    \includegraphics[width=1.0\linewidth]{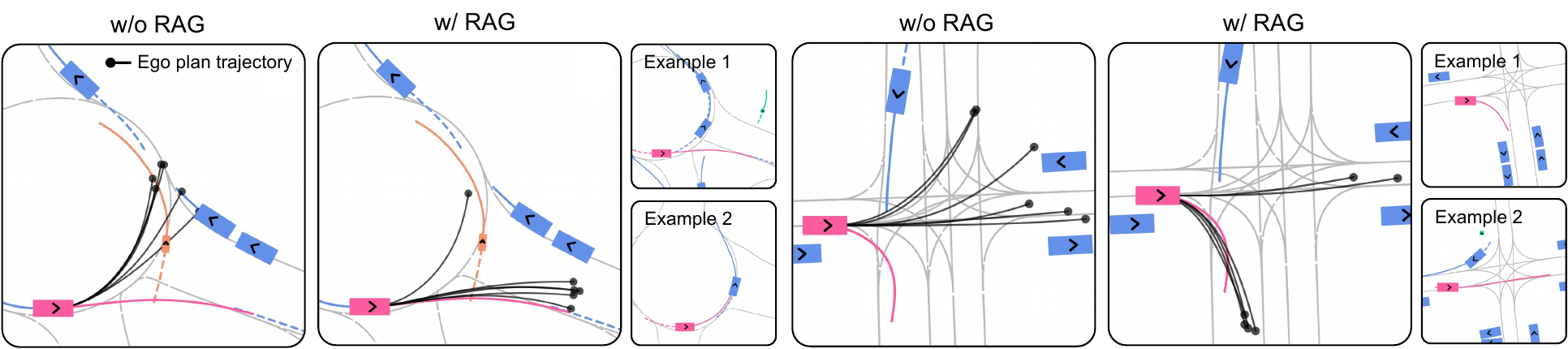}
    \vspace{-5mm}
    \caption{The comparison between the planners w/ RAG and w/o RAG. It is shown that using RAG avoids potential collision (left example), reduces out-of-road rate (right example), and increases planning trajectory diversity (both examples).}
    \label{fig:comparison}
    \vspace{-6mm}
\end{figure}

\textbf{Qualitative Examples.}
To gain further insight into how RAG influences planning behavior, we present two illustrative examples in Figure~\ref{fig:comparison}. In the left scenario, the planner without RAG generates a trajectory that risks collision with a nearby cyclist (orange) and fails to capture the multi-modal nature of the ground-truth trajectory. Augmentation with retrieved examples enables the planner to produce safer and more diverse trajectories. In the right scenario, the planner without RAG generates trajectories that deviate from the drivable area. With RAG, the generated plans are not only more accurate but also remain within valid driving boundaries, highlighting the effectiveness of retrieval-based augmentation in correcting critical planning errors.

\vspace{-3mm}
\subsection{Factors Influencing RealDrive Performance (\qa{Q3})}
\vspace{-1mm}

Given the numerous design components within RealDrive, we perform ablation studies to identify key factors that significantly affect its performance. Our analysis highlights four major considerations.

\textbf{Quality and Quantity of the Retrieval Database.}
A comparison between results on Tables~\ref{tab:nuscenes_results} and~\ref{tab:waymo_results} reveals that RAG yields greater improvements on Waymo. This discrepancy is largely attributable to the scale of the training sets: the nuScenes dataset contains only 700 training scenes, which limits the diversity of scenarios and reduces the diffusion model's capacity to extract informative features from retrievals. In contrast, the more extensive Waymo dataset enables better generalization and more effective retrieval. Furthermore, the use of oracle retrieval, i.e., segments from the exact same scene with access to ground-truth future trajectories, yields the highest performance gains. This observation underscores the potential for further improvements through the construction of a higher-quality and more comprehensive retrieval database.

\begin{wraptable}{R}{0.5\linewidth}
\vspace{-6mm}
\caption{Ablation study on Waymo dataset}
\label{tab:interpolation_type}
\centering
\resizebox{\linewidth}{!}{
\begin{tabular}{l|cccc}
    \toprule
    Ablation & w/o Obs. & w/o Act. & w/o TSE & Full \\
    \midrule
    minADE$_{6}$ ($\downarrow$) & 0.1604 & 0.1661 & 0.1674 & 0.1524 \\
    minFDE$_{6}$ ($\downarrow$) & 0.5753 & 0.6117 & 0.6232 & 0.5436 \\
    minCR ($\downarrow$)        & 0.0015 & 0.0016 & 0.0017 & 0.0012 \\
    avgCR ($\downarrow$)        & 0.0034 & 0.0033 & 0.0035 & 0.0026 \\
    minTTCE ($\uparrow$)        & 41.162 & 41.632 & 41.616 & 42.029 \\
    avgTTCE ($\uparrow$)        & 43.549 & 43.598 & 42.461 & 44.632 \\
    \bottomrule
\end{tabular}}
\vspace{-6mm}
\end{wraptable}

\textbf{Embedding Model.}
To evaluate the impact of the embedding model on retrieval quality and overall performance, we compare RealDrive against a variant that omits the task-specific embedding model (denoted as w/o TSE). This baseline adopts an encoder-decoder structure for learning scenario embeddings, following the approach in~\cite{ding2024realgen}. As shown in Table~\ref{tab:interpolation_type}, our task-specific embedding model, trained with a planning objective, outperforms the baseline, highlighting the importance of using downstream task representations for effective scenario retrieval.

\textbf{Interpolated Information.}
RealDrive performs interpolation on both observations and actions during the diffusion denoising process. To assess the contribution of each component, we conduct an ablation study comparing the effects of interpolating observation only, action only, and both. The results in Table~\ref{tab:interpolation_type} indicate that action interpolation is critical to the success of RAG in driving tasks. Nonetheless, combining observation and action interpolation yields the best performance, supporting our hypothesis that observation-level adaptation enhances the planner's ability to effectively leverage the retrieved action.

\begin{wrapfigure}{R}{0.35\linewidth}
    \centering
    \vspace{-7mm}
    \includegraphics[width=1.0\linewidth]{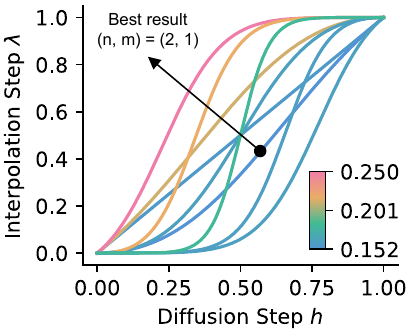}
    \vspace{-7mm}
    \caption{minADE$_6$ of different interpolation schedulers on Waymo. When the curve is flatter close to $h=0$, the performance is better.}
    \vspace{-4mm}
    \label{fig:sigmoid}
\end{wrapfigure}
\textbf{Design of the Interpolation Scheduler.}
An important design factor in RealDrive is the interpolation scheduler, which is defined by the sigmoid function in (\ref{equ:sigmoid}). This function governs the evolution of the interpolation coefficient $\lambda$ across the denoising steps $h$, with its shape controlled by two parameters. To evaluate its influence, we perform a parameter sweep and present the corresponding minADE$_6$ values in Figure~\ref{fig:sigmoid}, where different colors denote performance across various scheduler configurations.
Our analysis yields two key insights. First, scheduler curves that remain flatter near $h = 0$ are associated with improved planning performance. This suggests that interpolation should begin early, before significant denoising, so that the denoiser has sufficient time to adjust and refine the transferred action. Second, steeper scheduler curves tend to degrade performance, likely due to abrupt interpolation transitions that introduce significant deviations between consecutive states. These findings emphasize the importance of a gradual, well-paced interpolation schedule in enhancing retrieval-augmented planning.

\subsection{Impact of RAG on Inference Time (\qa{Q4})}

\begin{figure}[ht]
    \centering
    \vspace{-1mm}
    \includegraphics[width=1.0\linewidth]{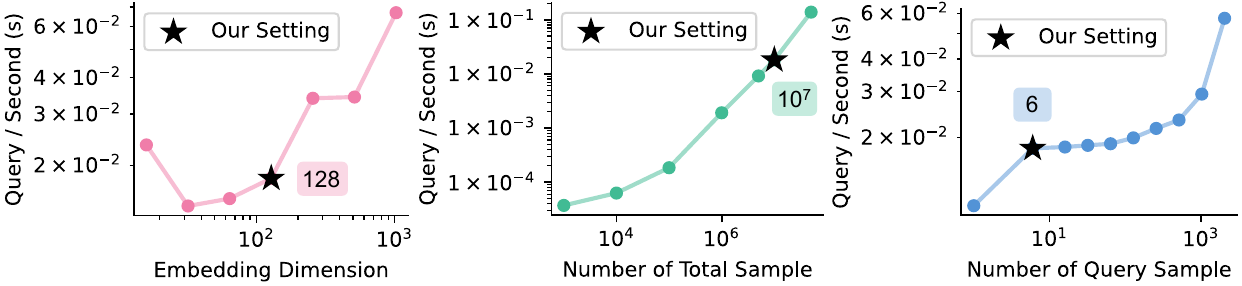}
    \vspace{-6mm}
    \caption{Time statistics of retrieving under different configurations. The setting we used for the Waymo dataset takes 0.0246 seconds.}
    \label{fig:speed}
    \vspace{-4mm}
\end{figure}

A critical requirement for driving planners is low-latency inference. However, incorporating retrieval during inference introduces additional computational overhead. To quantify this impact, we perform a detailed analysis of retrieval latency under various configurations, including embedding dimensionality, the size of the retrieval database, and the number of query samples.
Without retrieval, the base diffusion planner requires 0.0154 seconds per inference. When retrieval is incorporated, the total inference time increases to 0.0246 seconds. As illustrated in Figure~\ref{fig:speed}, under the configuration used in our setting on the Waymo dataset (embedding dimension of 128, $10^7$ retrieval entries, and 6 query samples), the retrieval step alone takes approximately 0.018 seconds per query. This latency is comparable to that of the planner itself, indicating that RAG can be feasibly integrated into real-time planning pipelines without introducing prohibitive delays.

\section{Conclusion and Limitation}
\label{sec:conclusion}

In this work, we present RealDrive, a novel Retrieval-Augmented Generation (RAG) framework for trajectory planning that combines task-specific retrieval with a diffusion model. Our approach addresses two key limitations of learning-based planners: limited generalization to rare, safety-critical scenarios and insufficient control over trajectory generation. By retrieving expert demonstrations relevant to the current driving context and interpolating them with the agent’s observations during the denoising process, RealDrive facilitates more reliable and controllable planning outcomes. Experiments conducted on two open-sourced datasets demonstrate that RealDrive substantially enhances trajectory diversity and generalization, reducing collision rates by up to 40$\%$ in long-tail scenarios.

Despite its demonstrated effectiveness, RealDrive exhibits two primary limitations. First, maintaining a large retrieval dataset imposes significant demands on on-device memory and can hinder real-time inference performance. Addressing this issue requires either selecting representative samples or compressing raw scenarios into structured memory. Second, the current implementation of RealDrive operates on vectorized inputs. Future research will focus on extending the framework to accommodate video input, enabling support for fully end-to-end autonomous driving systems.

\clearpage
\newpage
\bibliographystyle{unsrt}
\bibliography{reference}

\newpage
\appendix

\section{Potential Societal Impacts}
\label{sec:impact}

While RealDrive improves planning performance and safety by leveraging retrieved expert demonstrations, it also introduces risks from reliance on retrieval-based priors. In particular, if the retrieval database contains examples that encode undesirable or unsafe behaviors -- either unintentionally or through adversarial poisoning -- the planner may generate risky trajectories that mirror those examples. As a mitigation strategy, it is essential to implement rigorous filtering and validation procedures to ensure that retrieved trajectories align with safety constraints and traffic rules. Furthermore, interpretability tools and uncertainty estimation should be employed to monitor and audit the influence of retrieved samples during deployment.

\section{More Experiment Results}

\subsection{Qualitative examples of retrieval model}

\begin{figure}[ht]
    \centering
    \vspace{-2mm}
    \includegraphics[width=1.0\linewidth]{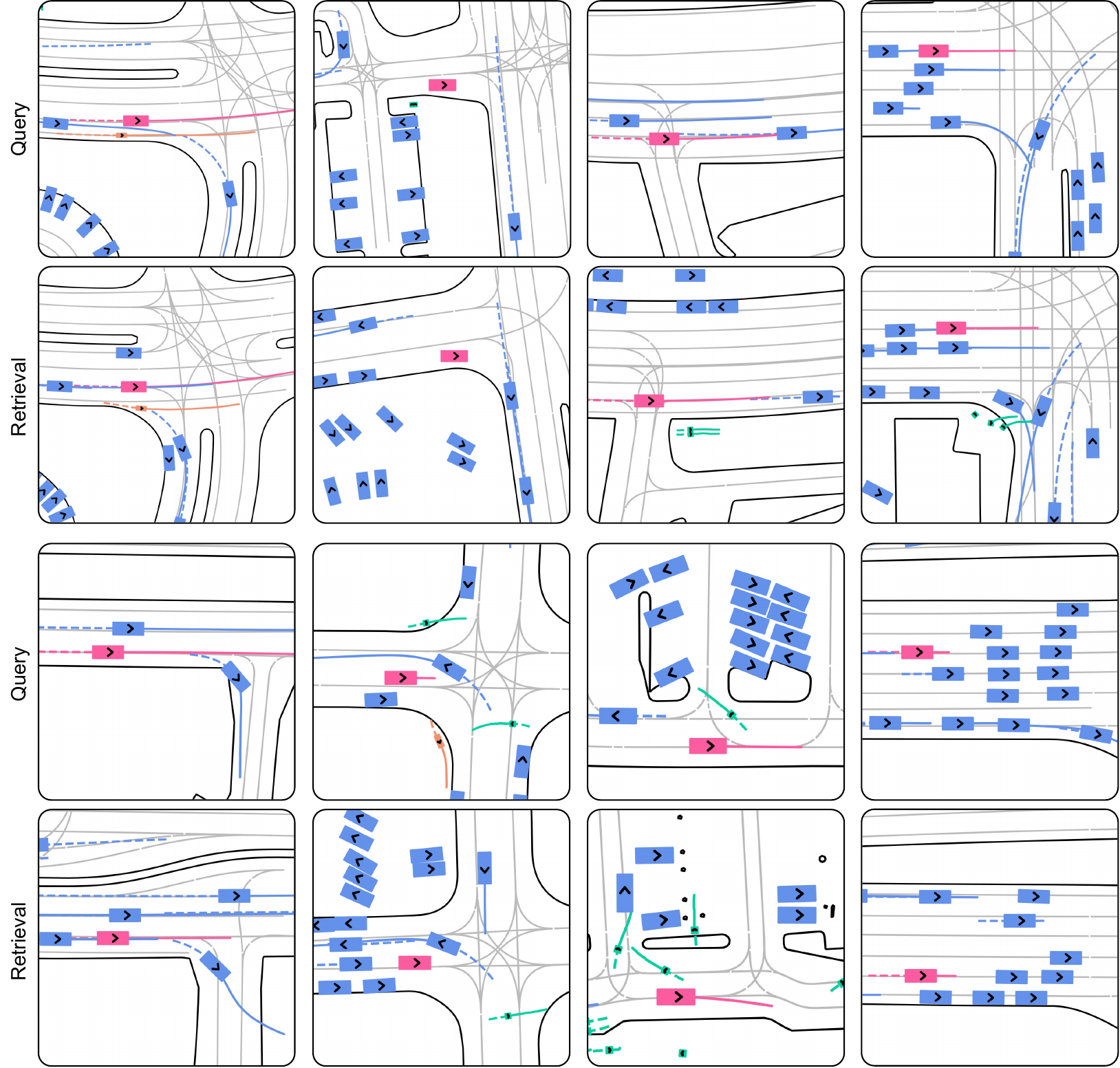}
    \vspace{-6mm}
    \caption{More examples of the retrieval model.}
    \label{fig:more_retrieval}
    \vspace{-4mm}
\end{figure}

We provide more examples to demonstrate the quality of our retrieval embedding model in Figure~\ref{fig:more_retrieval}.

\subsection{Comparison between w/ RAG and w/o RAG models on long-tail scenarios.}

\begin{figure}[ht]
    \centering
    \includegraphics[width=1.0\linewidth]{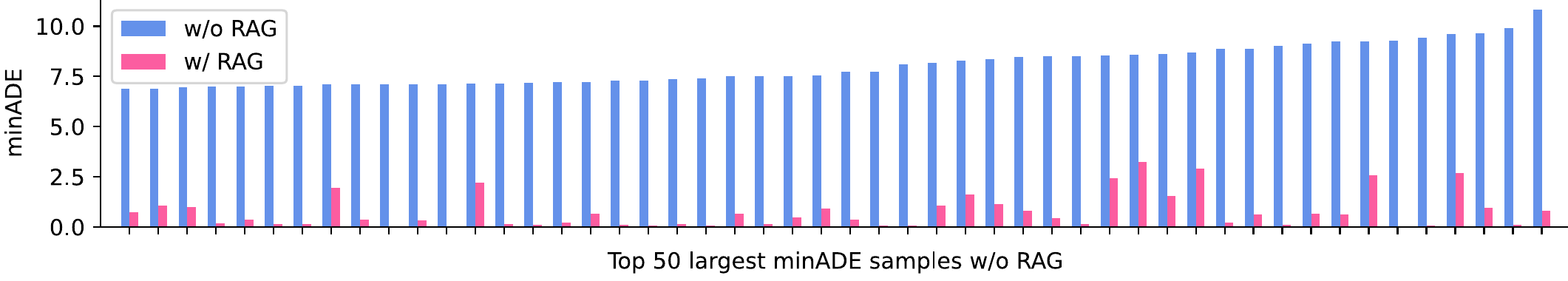}
    \vspace{-6mm}
    \caption{Top 50 samples with largest minADE of w/o RAG model.}
    \label{fig:top50}
    \vspace{-4mm}
\end{figure}

We find the top 50 samples according to the minADE of the w/o RAG model and plot them in Figure~\ref{fig:top50}. We also plot the minADE results for the same samples using the w/ RAG model. We find that using RAG significantly reduces the minADE.

\subsection{Qualitative examples of model outputs}

\begin{figure}[ht]
    \centering
    \vspace{-2mm}
    \includegraphics[width=1.0\linewidth]{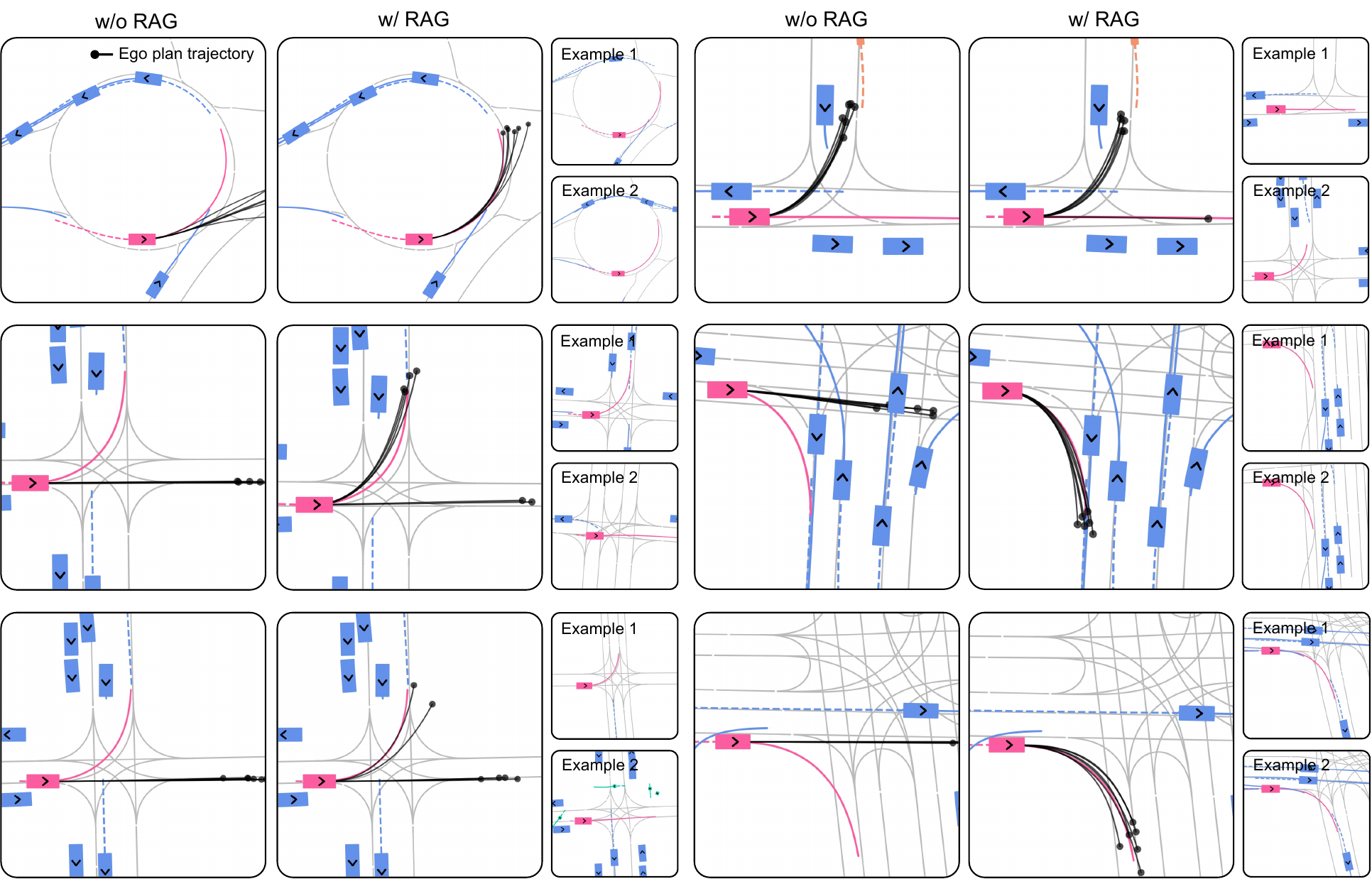}
    \vspace{-6mm}
    \caption{More examples of the comparison between different model outputs.}
    \vspace{-5mm}
    \label{fig:more_comparison}
\end{figure}

We provide more examples that compare the trajectory outputs from models without RAG and with RAG in Figure~\ref{fig:more_comparison}. The model using RAG generates more diverse and safe planning trajectories.

\section{Experiment Details}

\subsection{Details of Metric}
\label{app:metric}

The minADE$_k$ and minFDE$_k$ metrics are defined by the following equation:
\begin{equation}
    \text{minADE}_k = \frac{1}{NT_F} \min_k \sum_{n=1}^N \sum_{t=1}^{T_F} \|\hat{x}_{n,t} - x^k_{n,t} \|_2,
\end{equation}
\begin{equation}
    \text{minFDE}_k = \frac{1}{N} \min_k \sum_{n=1}^N \|\hat{x}_{n,T_F} - x^k_{n,T_F} \|_2,
\end{equation}
where $x^k_{n,t}$ is the $k$-th sample for the agent $n$ at timestep $t$.
The minimal collision rate (CR) and average collision rate metrics are defined by:
\begin{equation}
\mathrm{minCR} = \min_k
\begin{cases}
1, & \text{if}\ \ \sum_{n=2}^{N}\sum_{t=1}^{T_F} \text{Collision}(x_{1, t}, x_{n, t}) > 0, \\
0, & \text{otherwise},
\end{cases}
\end{equation}
\begin{equation}
\mathrm{avgCR} = \sum_{k=1}^K
\begin{cases}
1, & \text{if}\ \ \sum_{n=2}^{N}\sum_{t=1}^{T_F} \text{Collision}(x_{1, t}, x_{n, t}) > 0, \\
0, & \text{otherwise},
\end{cases}
\end{equation}
where the function $\text{Collision}(x_{1, t}, x_{n, t})$ outputs 1 if the distance between two vehicles are smaller than a threshold.
The Time to closest encounter (TTCE) metric we used in the experiments is calculated by:
\begin{equation}
    TTCE = \min\left( \max\left( \dfrac{(\mathbf{p} \cdot \mathbf{v})}{\|\mathbf{v}\|^2 + \delta},\, 0 \right),\; \tau_{\text{max}} \right),
\end{equation}
where $\mathbf{p} = \mathbf{x}_{\text{agent}} - \mathbf{x}_{\text{ego}}$ is the relative position vector, $\mathbf{v} = \mathbf{v}_{\text{agent}} - \mathbf{v}_{\text{ego}}$ is the relative velocity vector, $\delta$ is a small positive constant $10^{-8}$) to ensure numerical stability, and $\tau_{\text{max}}$ is the maximum allowed TTCE, which is set to 4 in our experiments.

\subsection{Forward Dyanmics Model}
\label{app:dynamics}

We consider a bicycle model as our forward dynamics, which has the following transition function:
\begin{equation}
\mathcal{F}(a_t) = \mathcal{F}(u_t, \delta_t) = 
\left\{
\begin{aligned}
x_{t+1} &= x_t + h_t \cos(\theta_t) \cdot \Delta t \\
y_{t+1} &= y_t + h_t \sin(\theta_t) \cdot \Delta t \\
h_{t+1} &= h_t + u_t \cdot \Delta t \\
\theta_{t+1} &= \theta_t + \frac{h_t}{L} \tan(\delta_t) \cdot \Delta t
\end{aligned}
\right.
\end{equation}
where $L$ is the wheelbase of the vehicle in meter, $\delta$ is the steering angle of the front wheels, and $u$ is the longitudinal acceleration.

\subsection{Compute Resources}
\label{app:compute}

Experiments are conducted on a desktop with AMD Ryzen 9 7950X 16-Core Processor, NVIDIA RTX 6000 Ada Generation GPU with 48GB memory, and 128GB memory.
The training on the Waymo dataset takes about 72 hours.

\subsection{Hyperparameter}
\label{app:hyperparameter}

We summarize all hyperparameters used in the experiments in Table~\ref{tab:hyperparameter}. Note that we use identical hyperparameters for the nuScenes and Waymo datasets, except for the learning rate scheduler and batch size. For the RAG part, we only introduce two additional hyperparameters $m$ and $n$.

\begin{table}[ht]
\caption{Hyperparameters of experiments}
\label{tab:hyperparameter}
\vspace{-1mm}
\centering{\resizebox{1.0\linewidth}{!}{
\begin{tabular}{c|l|c|c}
    \toprule
    Notation & Description & Value (nuScenes) & Value (Waymo) \\
    \midrule
    $\Delta_t$  & time interval between two waypoints in the trajectory & $0.1$ & $0.1$ \\
                & step size of segment window                           & $0.1$ & $0.1$ \\
    $T_h$       & time length of history (second)                       & $2$ & $2$ \\
    $T_f$       & time length of future (second)                        & $4$ & $4$ \\
    $A$         & max number of surrounding agents in the scene         & $20$ & $20$ \\
    $S$         & max number of lane in the map                         & $100$  & $100$ \\
    $P$         & number of points per lane                             & $50$  & $50$ \\
    \midrule
    $T_{max}$   & total training steps                                  & $2\times 10^4$ & $2\times 10^5$ \\
    $B$         & batch size                                            & $64$ & $128$ \\
    $lr$        & initial learning rate                                 & $5\times e^{-4}$ & $5\times e^{-4}$ \\
                & multi-step learning rate scheduler                    & $[3, 6, 9, 12, 15]\times 10^3$ & $[2, 4, 6, 8, 10]\times 10^4$ \\
                & learning rate scheduler decay discount                & $0.5$ & $0.5$ \\
                & weight decay of AdamW optimizer                       & $0.01$ & $0.01$ \\
                & max gradient clip value                               & $5$ & $5$ \\
                & discount of state loss over time steps                & $0.95$ & $0.95$ \\
    \midrule
    $H$         & diffusion step                                        & $10$ & $10$ \\
    $h_e$       & dimension of hidden embedding                         & $128$ & $128$ \\
    $h_{ff}$    & dimension of the feedforward network in attention     & $512$ & $512$ \\
    $h_{n}$     & number of head in attention                           & $8$ & $8$ \\
                & number of encoder layer                               & $1$ & $1$ \\
                & number of decoder layer                               & $2$ & $2$ \\
                & dropout rate                                          & $0.1$ & $0.1$ \\
    \midrule
    $(n, m)$    & parameter of the interpolation scheduler              & $(2, 1)$ & $(2, 1)$ \\
    \bottomrule
\end{tabular}}}
\end{table}



\end{document}